\documentclass{article}
\usepackage{arxiv}

\usepackage[utf8]{inputenc} % allow utf-8 input
\usepackage[T1]{fontenc}    % use 8-bit T1 fonts
\usepackage[bookmarks=false]{hyperref}      % hyperlinks
\usepackage{url}            % simple URL typesetting
\usepackage{booktabs}       % professional-quality tables
\usepackage{amsfonts}       % blackboard math symbols
\usepackage{nicefrac}       % compact symbols for 1/2, etc.
\usepackage{microtype}      % microtypography
\usepackage{graphicx}
\graphicspath{{Pics/}}

\usepackage[linesnumbered,ruled]{algorithm2e}
\usepackage{amsmath}
\DeclareMathOperator*{\argmax}{arg\,max}

\title{ELMV: an Ensemble-Learning Approach for Analyzing Electrical Health Records with Significant Missing Values}

\author{
\begin{tabular}{cc}
Lucas Jing Liu\footnotemark[1]\thanks{Equal contributors: Lucas Jing Liu and Hongwei Zhang.}     & Hongwei Zhang\footnotemark[1]  \\
Department of Computer Science               & Department of Metabolic \& Bariatric Surgery \\
University of Kentucky                       & Shanghai Jiaotong University                 \\
Lexington, KY                                & Affiliated 6th People’s Hospital             \\
jli394@uky.edu                               & Shanghai, China                              \\
                                             & zhw00sub@163.com                             \\
                                             &                                              \\
Jianzhong Di                                 & Jin Chen                                     \\
Department of Metabolic \& Bariatric Surgery & Department of Internal Medicine              \\
Shanghai Jiaotong University                 & Department of Computer Science               \\
Affiliated 6th People’s Hospital             & University of Kentucky                       \\
Shanghai, China                              & Lexington, KY                                \\
dijianzhong@sjtu.edu.cn                      & chen.jin@uky.edu                            
\end{tabular}
}

\begin{document}
\maketitle

\begin{abstract}
 Many real-world Electronic Health Record (EHR) data contains a large proportion of missing values. Leaving substantial portion of missing information unaddressed usually causes significant bias, which leads to invalid conclusion to be drawn. On the other hand, training a machine learning model with a much smaller nearly-complete subset can drastically impact the reliability and accuracy of model inference. 
 Data imputation algorithms that attempt to replace missing data with meaningful values inevitably increase the variability of effect estimates with increased missingness, making it unreliable for hypothesis validation.
 We propose a novel Ensemble-Learning for Missing Value (ELMV) framework, which introduces an effective approach to construct multiple subsets of the original EHR data with a much lower missing rate, as well as mobilizing a dedicated support set for the ensemble learning in the purpose of reducing the bias caused by substantial missing values. ELMV has been evaluated on a real-world healthcare data for critical feature identification as well as a batch of simulation data with different missing rates for outcome prediction. On both experiments, ELMV clearly outperforms conventional missing value imputation methods and ensemble learning models. %The results indicate that ELMV is able to identify critical features and predict outcomes in the presence of large missingness in EHR data. 
\end{abstract}


\section{Introduction}
Real-world Electronic Health Record (EHR) data have played an important role in improving patient care and clinician experience and providing rich information for biomedical researches~\cite{goetz2012ehrs,weng2012using,shickel2017deep}. However, many EHR data contain a significant proportion of missing values, which could be as high as 50\%, leading to a substantially reduced sample size even in initially large cohorts if we restrict the analysis to individuals with complete data~\cite{molenberghs2007missing,ibrahim2012missing}. On the other hand, leaving a big portion of missing information unaddressed usually cause bias, loss of efficiency, and finally leads to inappropriate conclusion to be drawn~\cite{little2019statistical}. %In particular, training a machine learning or statistical predictive model on a complete but much small subset of the original data or EHR data with either small data without missingness or original data that have a lot of missing values can both drastically impact the reliability and accuracy of model inference [XX]. 

Data imputation algorithms (e.g. the scikit-learn estimators~\cite{pedregosa2011scikit}) attempt to replace missing data with meaningful values including random values, the mean or median of rows or columns, spatial-temporal regressed values, most frequent values in the same columns, or representative values identified using k-nearest neighbor~\cite{efron1994missing}. Advanced data imputation algorithms, such as Multivariate Imputation by Chained Equation (MICE)~\cite{JSSv045i03}, have been developed to fill missing values multiple times. Leveraging the power of GPU and big dta, deep neural network models, such as Datawig~\cite{biessmann2019datawig}, can estimate more accurate results than traditional data imputation methods ~\cite{beaulieu2017missing}. However, as stated in the statistical literature~\cite{mcneish2017missing,clavel2014missing,mishra2014comparative}, there is inevitably increasing variability of effect estimates with increased missingness, and results may not be reliable enough for hypothesis validation if more than 40$\%$ data are missing in important variables~\cite{dong2013principled,jakobsen2017and}, indicating that data imputation is not a go-to solution when a significant portion of the values is missing. 
Furthermore, missing data in clinical studies do not occur at random. Certain data points are missing because of patient drop out, treatment toxicity, or biomarkers that are difficult to measure~\cite{ibrahim2012missing}. Applying data imputation algorithms designed for missing-at-random to EHR data may lead to biases in model prediction~\cite{li2018don}. 

%\textcolor{blue}{It is true, but in our design, we do not test out simulation data that are missing not at random ,we only simulated missing at random dataset, so if we add these here, I wonder if people will criticize the usefulness of ELMV for dataset missing not at random. }\textcolor{red}{This is a common problem and reviewers will ask it sooner or later. We add these here, and address this issue using the real data.}

On the real-world EHR data, a inference models that account for the missing data must consider the reasons for missingness~\cite{madley2019proportion}. 
We observe that in the EHR data, important variables are likely to be retained by auxiliary variables. For example, hemoglobin A1c (HbA1c) is an important index for diabetes patients. By measuring HbA1c, clinicians can get an overall picture of the average blood sugar levels over a few months. Multiple clinical measurements, such as fasting blood glucose, are highly correlated with HbA1c~\cite{ghazanfari2010comparison} and are often found in the EHR of diabetes studies.
Hence, if HbA1c is missing, a well trained predictive model can still rely on the auxiliary features of HbA1c, thus maintaining a relatively high performance. 

\begin{figure*}[bt!]
  \centering
  \includegraphics[width=0.8\linewidth, height=10cm]{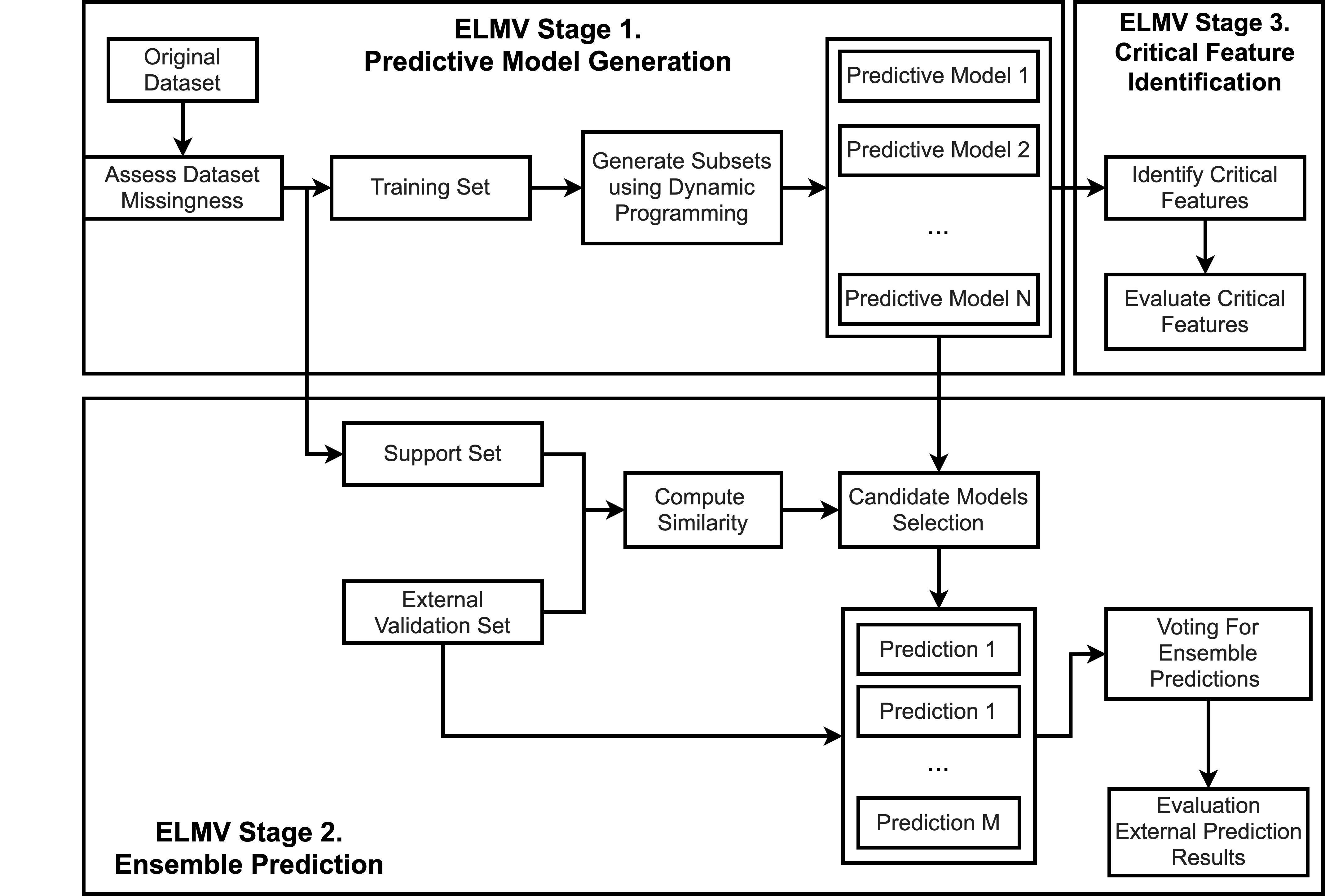}
  \caption{Overall framework of ELMV. It includes three stages: predictive model generation, ensemble prediction, and critical feature identification.}
\label{general_f}
\end{figure*}

In this project, we present a novel method called Ensemble-Learning for Missing Value (ELMV) to analyze EHR data with significant missing values, aiming to identify unbiased precise predictive patterns from EHR data. Specifically, given an EHR dataset with a significant missing rate, ELMV first generates multiple qualified maximal subsets of the original EHR data using dynamic programming. These qualified maximal subsets have much smaller missing rates than the original data. And then, ELMV trains predictive models using every qualified maximal subset and save the trained model for further use. Finally, for each record in the external validation data, ELMV selects multiple pre-trained models and employs ensemble learning for the final prediction. %
The architecture of ELMV is illustrated in Figure~\ref{general_f}. 
Experimental results on a real-world healthcare data as well as a batch of simulation data with different missing rates show that ELMV clearly outperforms conventional missing value imputation methods and traditional ensemble learning models.

Overall, ELMV has four algorithmic advantages:
\begin{itemize}
    \item To our knowledge, ELMV is the first ensemble learning approach that is capable of analyzing large EHR data with significant missing values accurately without using data imputation. 
    \item By constructing multiple maximal subsets of the original EHR data, opportunities are that even if critical features are removed due to high missingness, predictive models built using auxiliary features may still maintain a relatively high performance. 
    \item ELMV introduces a a dedicated support set for the ensemble learning in the purpose of reducing the bias caused by substantial missing values. 
    \item ELMV can identify critical features from large EHR data with significant missing values. 
\end{itemize}

%This study focuses on identifying critical features for predicting the patterns of patients’ outcome trajectory when there are substantial missingness of features. The experiment was done on diabetes patient 3-yr follow-up data for predicting HbA1c outcome trajectory.  We successfully identify 10 critical features for distinguish patients with different HbA1c trajectory and our best model achieved the best Accuracy of \textcolor{red}{XXX} , which outperforms other baseline models.  The experimental results show that  ELMV  is able to capture clinical factors that are known to be associated HbA1c trajectory in the presence of large missingness in the data.

\section{Background}
In the recent years, many techniques have been developed to handling missing values in big data. Among them, the simplest and most prevalent strategy is to conduct complete-cases analysis (CCA), which refers to removing records with any missing values and focus only on patients who have the complete records of all parameters~\cite{wells2013strategies}. In practice, removing patients with any missing values will inevitably introduce biases given that there is often a huge difference between the true distribution of all patients and that of the patients with complete records~\cite{pigott2001review}. In addition, regarding inference model training, the CCA strategy will significantly reduce the training size, resulting in models under-trained.  

Another common strategy for handling missing values is data imputation. Imputation techniques can be categorized into two groups: single imputation and multiple imputation~\cite{donders2006gentle}. The single imputation refers to replacing a missing value with an estimated value~\cite{blankers2010missing}. An example of the simple imputation strategy is the mean imputation~\cite{hu2017strategies}, where a missing value is replaced with the arithmetic mean. The problem of the simple imputation strategy is that it may significantly underestimate the variance of the data and ignores the complex relationships among explanatory variables~\cite{wells2013strategies}. This problem can be addressed using more sophisticated single imputation methods such as regression imputation and the expectation-maximization (EM) algorithm, in which a missing value is assigned by studying the statistical relationships between the target variable and the rest variables in the same dataset~\cite{hu2017strategies}. 
In contrast, multiple imputation techniques estimate a missing value with multiple imputed data. One such technique is Multivariate Imputation by Chained Equations (MICE), where the statistical uncertainty of different imputed data is taken into account~\cite{azur2011multiple}. However, none of the existing imputation method outperforms the others on every data, indicating that there is no universal model~\cite{hu2017strategies} for missing value imputation. 

As a matter of fact, most machine learning models can only be applied on complete data or will automatically conduct a complete-cases analysis~\cite{hu2017strategies}.  
However, XGBoost~\cite{chen2016xgboost}, a recent implementation of the gradient boosting model, can automatically handle missing values with its built-in mechanisms. Specifically, XGBoost handles the missing data problem by adding a default direction for missing values in each tree splitting. The optimal direction for a missing value in each particular explanatory variable at each tree node is learned during model training  with the aim to minimize the regulated loss~\cite{chen2016xgboost}. 
%
%If there is no missing value in a particular explanatory variable in training stage, but the external validation set has missing values in that particular variable, the XGBoost model choose the default direction. 
%
A potential problem in handling missing value in XGBoost is that XGBoost will always choose the default direction for model prediction on the validation set. Thus, the prediction could be a random guess if the missingness patterns in training and validation are entirely different. This could be the case when large amount of missing value existing in the data.  

Overall, the common problem that existing approaches have is that they do not have the adaptability for handling large missing values. In addition, the discrepancy between training and validation has not been well addressed regarding model inference. To address these problems, we propose ELMV, an ensemble learning framework that 1) is capable of handling substantial missing values, 2) has the adaptability to different dataset and different predictive models; and 3) takes into account the discrepancy between training and validation in terms of both missingness and pattern recognition.

\section{Method}
ELMV aims to identify unbiased precise predictive patterns from EHR data, which if learned directly, may result in biases caused by substantial missing values \cite{beaulieu2018characterizing}. Specifically, given an EHR dataset $I$ with significant missing values, ELMV first generates a set of subsets of $I$ with small missing rates, denoted as $S$, using dynamic programming, and upon these, builds predictive models $M$ so as to mitigating the overall bias in each dataset in $S$ for a single predictive model. Second, for every record in the external validation data, ELMV selects the most suitable models from $M$ for the final prediction using ensemble learning. %, meaning that the final prediction is made by aggregating the information learned from candidate classifiers by proposed strategies into one final consensus result.

Since ELMV is a general machine learning framework for learning from EHR data with significant missing rates, any conventional machine learning model, such as XGBoost ~\cite{chen2016xgboost} and SVM ~\cite{noble2006support}, can be used in our framework. For the demonstration purpose, we used XGBoost in the paper. %In the following subsections, we explain the procedures in each stage.

The framework of ELMV involves three stages namely model generation, critical feature identification, and outcome prediction.  The architecture of ELMV is illustrated in Figure~\ref{general_f}.

\subsection{ELMV Stage 1. Predictive Model Generation}
In the predictive model generation stage, we first compute the data missingness of a given EHR dataset, assessing whether it is appropriate to use ELMV. And then, we generate multiple subsets of the original data with low missing rates. Finally, a predictive model is trained on each subset.

\subsubsection{Assessing Data Missingness}\label{sec:Missingness}
Given an EHR dataset $I$ where rows are patients $P$, columns are features $F$ in the EHR, $N_p$ is the total number of patients, $N_f$ is the total number of features, a missingness indicator for each patient $p$, denoted as $MissingI^{p}$, is defined as a binary vector with the length of $N_f$ where a one in a specific entry represents that the corresponding feature is missing for patient $p$. 

Specifically, for EHR data with temporal features $TF^{N_f,T,N_p}$, where $T$ is the number of time points of a temporal feature and $TF\subseteq T$, we define the missingness indicator as a two-dimensional matrix : for each temporal feature of patient $p$, denoted as $tf_j^p \in TF$, if a temporal trend-based feature is missing because of the missing data at time point $t_i$, let $MissingI^{p}(t_i,tf_j)= 1.$.

A 2-dimensional binary matrix, denoted as $MissingI \in \mathbb{R}^{N_p \times N_f}$ can then be generated to store the missingness information of all patients. In 2D case,  $MissingIM[i,j] = 1 $ representing the patient $i$ has a missing value in the $j^{th}$ feature. In the case of 3D temporary data set where the third dimension represents time of records, $MissingIM[i,j] = 1 $ representing the patient $i$ at least have one data point missing in the time trajectory of the $j^{th}$ feature.

Based the definition of data missingness, we compute the missing rate of the entire dataset $I$, assessing whether ELMV or data imputation techniques should be used. Typically, if the data missingness is low, it is appropriate to impute missing data. However, if the missing rate is above 40\%, data imputation may inevitably increase the variability of effect estimates. Instead of imputing missing values directly, ELMV relies on ensemble learning which aggregates predictive models built on multiple subsets with significantly smaller missing rates. 

Note that although ELMV is still applicable when the missing rate is low (e.g. under 10\%), its performance is similar to the other state-of-art models. 

\subsubsection{Generating Subsets with Low Data Missingness} 
Given an EHR dataset $I$ and a user-defined data missing rate up-bound $T_{max-missing}$, e.g. 20\%, which is much lower than the missing rate of $I$, we generate a set of maximal subsets of $I$ with its missing rate lower than or equal to $T_{max-missing}$, saved in $S$. %, using dynamic programming.
A subset $s$ of $I$ is a 2-dimensional matrix where rows are patients and columns are features in EHR. We say $s$ is maximal if and only if its missing rate can only increase if its rows or columns are replaced by any new rows or columns in $I$. 

Since the total number of possible subsets is ${N_p \choose x} \times { N_f \choose y }$, where where $x$ and $y$ are the numbers of rows and columns of $s$, it is clearly impractical to enumerate all the possibilities and then select the maximal ones. Thus, to identify all the qualified maximal subsets of $I$ with missing rates lower than or equal to $T_{max-missing}$, we developed a two-step approach.  

The approach for generating qualified maximal subsets consists of two steps: 1) to generate all the maximal subsets using dynamic programming, and 2) to filter the maximal subsets that have nearly duplicated information. The pseudocode for maximal subsets generation is illustrated in Algorithm~\ref{DP_alg}. In the following section, we explain the steps for generating the qualified maximal subsets.

%\textcolor{blue}{NOTe: The algorithm only for maximal subsets, does not include the filtering part.}

In the first step, we track the missingness of all the subsets-to-generate using a 2-dimensional matrix $MissingC \in \mathbb{R}^{N_p \times N_f}$. The value at each entry $MissingC(x,y)$ represents the minimum number of missing values of any subset of $I$ with $x$ patients and $y$ features. For instance, $MissingC(100, 200) = 1300$ means that the minimum number of missing values is 1300 for any sub-matrix of $I$ with 100 patients and 200 features. $MissingC$ can be used to select maximal subsets (see details in Algorithm~\ref{DP_alg}). 

In the dynamic programming process, we start to fill $MissingC$ and to generate the corresponding maximal subsets from the bottom right corner, $MissingC(N_p,N_f)$. Naturally, it represents the number of missing values when all features and all patients are selected. Hence, the corresponding maximal subset is $I$ itself. 
And then, we repeatedly remove one feature or one patient that has the maximum number of missing values at a time until the subset reaches the smallest required number of features and the smallest number of patients. 
By removing a feature or a patient that has the maximum missing values at each time step, the generated subset is ensured to have the minimum missing ratecorresponding to the required number of features and patients. The whole process is achieved using dynamic programming~\cite{Bellman34}.  

The second step of subset generation is to identify and remove subsets conveying nearly identical information. %and keeping the subsets that conveys the maximum amount of information within the same missing percentage category. 
For all the subset with a similar missing rate, we keep the subsets with the maximum number of features if the number of patients is identical, or keep the subsets with the maximum number of patients if the number of features is identical. %Otherwise, we keep XXX. 
%\\
%\textcolor{blue}{Comments: I am not sure about this otherwise part, and in the first sentence, it is not similar missing rate, it is nearly the same missing rate, it is not between $5\%$ to $10\%$, it is either close $5\%$ or close to $1\%$ or equals $0\%$ -- Lucas}

The final outcome of this step is a set of maximal subsets of the original EHR dataset with missing ratio smaller than or equal to a user-defined data missing rate up-bound $T_{max-missing}$.

\begin{algorithm}
\LinesNotNumbered
\SetAlgoLined
\SetKwInOut{Input}{Input}
\SetKwInOut{Output}{Output}
\SetKwInOut{Intermediate}{Intermediate}
\Input{$2D\:DataMatrix [N_p \times N_f]$ or $3D\:Temporal\:DataMatrix\:[N_p \times N_f \times N_t]$ } 
\Intermediate{$MissingI,\;MissingI\_List,\:MissingC,$}
\Output{$Max\_S$  \# Maximal Subsets}

  \DontPrintSemicolon
  \SetKwFunction{Fone}{ConstructMissingI}
  \SetKwProg{Fn}{Function}{:}{}
  \Fn{\Fone{$DataMatrix$}}{
        \For{$i=N_p$ \KwTo $1$}{
            \For{$j=N_f$ \KwTo $1$}{
                \uIf{$\sum_{t=1}^{N_t} MissingI^{p}[t,j] \geq 1$}{
                 $MissingI_{i,j} = 1$ 
                }
                \Else{
                 $MissingI_{i,j} = 0$
               }
            }
        }
   \KwRet{$MissingI$}
  }
  \;
  
    \DontPrintSemicolon
  \SetKwFunction{Ftwo}{Order}
  \SetKwProg{Fn}{Function}{:}{}
  \Fn{\Ftwo{$MissingI$}}{
     Order input by the missing percentage of patients and features ascendingly from left to right and from top to bottom
     
   \KwRet{ordered $MissingI$}
  }
  \;
  
  \DontPrintSemicolon
  \SetKwFunction{Fthree}{CountMissings}
  \SetKwProg{Fn}{Function}{:}{}
  \Fn{\Fthree{$MissingI$}}{
     Count the total number of ones in input \;
   \KwRet{Total\_Number\_Of\_Missing\_Values}
  }
  \;
  
Initialization\;
$MissingI\_List_{N_p,N_f}$ = \Fone{$DataMatrix$} \;
$MissingC_{N_p,N_f}$ = \Fthree{$MissingI\_List_{N_p,N_f}$}; \;
$MissingI\_List_{N_p,N_f}$ = \Ftwo{$MissingI\_List_{N_p,N_f}$} \;

\For{$i=N_p $ \KwTo $1$}{
    \For{$j=N_f$ \KwTo $1$}{
    \uIf{$i \; != N_p \; and \; j \; != N_f$}{
        \uIf{$MissingC_{i,j+1} < MissingC_{i+1,j}$ or $MissingC_{i+1,j}$ is empty}{
            $MissingI\_List_{i,j+1}$ = \Ftwo{$MissingI\_List_{i,j+1}$} \;
             $last\_step$ = $MissingI\_List_{i,j+1}$\;
             \tcc{then remove the last feature}
             $MissingI\_List_{i,j}$ =  $last\_step[ , -last\:column]$ \;
        }
        \uElseIf{$MissingC_{i,j+1} \geq MissingC_{i+1,j}$ or $MissingC_{i,j+1}$ is empty}{
             $MissingI\_List_{i+1,j}$ = \Ftwo{$MissingI\_List_{i+1,j}$} \;
             $last\_step$ = $MissingI\_List_{i+1,j}$\;
             \tcc{then remove the last patient}
             $MissingI\_List_{i,j}$ =  $last\_step[ , -last\:row]$\;      
        }
       $MissingC_{i,j}$ = \Fthree{$MissingI\_List_{i,j}$}\;
        $Max\_{S_{i,j}}$ = Patient and Features in $MissingC_{i,j}$\;
    
    }
    
    }
}

\caption{Algorithm For Generating Maximal Subsets}
\label{DP_alg}
\end{algorithm}

\subsubsection{Training Predictive Models}
Using every qualified maximal subset of the original data $I$, we train a traditional classification model and save all the trained models in model set $M$. 
Since ELMV is a general framework for learning predictive patterns from data with significant missingness, any classification model, such as support vector machine and gradient boosting, can be used in this step. We expect that the classification model deployed here is capable of handling a few missing values. Otherwise, we recommend to employ a data imputation method before calling a classification model. 
%
%Any parameter settings of XGBoost and any sampling method can be selected according to the needs of different classification tasks. 
%

For the demonstration purpose, the XGBoost implementation \cite{chen2016xgboost} ''xgboost'' in R library is used in this step. Specifically, we choose a tree-based model called ``gbtree'' booster for relatively easier classification task with a softmax objective ''multi:softprob''. Also, we choose a linear model called ''gblinear'' for relative harder classification task with a logistic objective ``binary:logistic'', such that the multi-class task can be converted into binary classification using the one vs. rest approach~\cite{aly2005survey}. Finally, each trained predictive model is evaluated using either 10-fold cross-validation or the leave-one-out validation approach. Model performance is saved in $M$ for later use.

%\textcolor{red}{section above updated -- Lucas}
%{\bf instead of listing ``gbtree'' or ``gblinear'' boosters and their objective functions, we need to be self-contained. it means we need to describe what is gbtree or gblinear briefly. -- JIN}

\begin{figure*}[!bt]
\centering
\includegraphics[width=0.8\linewidth]{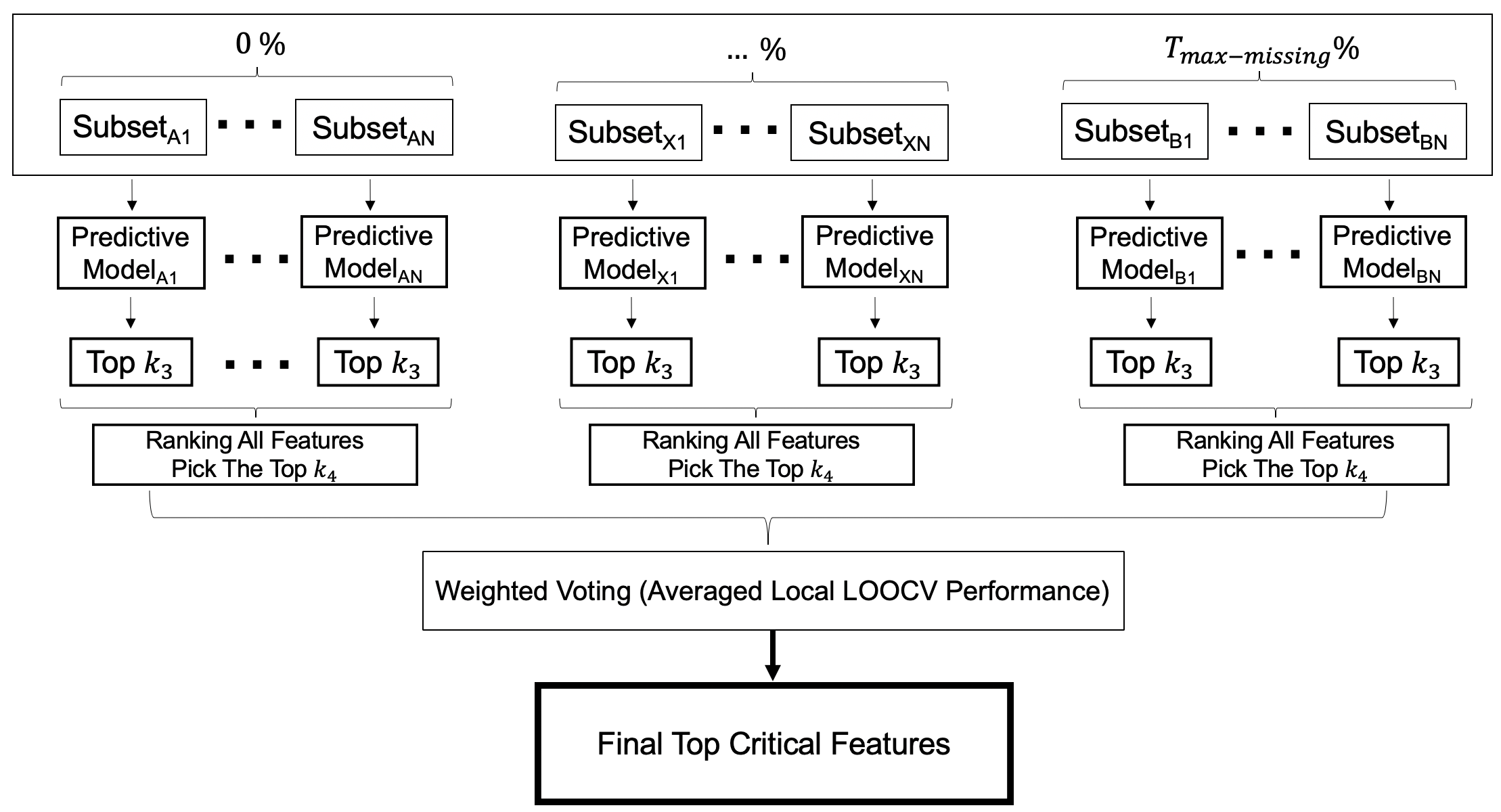}
\caption{The framework of critical feature identification using ELMV.}
\label{feature_p}
\end{figure*}

\subsection{ELMV Stage 2. Ensemble Prediction}
In the ensemble prediction stage, ELMV aggregates multiple selected predictive models trained in stage one to make predictions for records in an external validation set. 
%\textcolor{blue}{Comments: Can we use external validation set? We also had it in the process figure, or I can make the corresponding changes in the figure} \textcolor{red}{okay let's call it external validation set. Please change the content accordingly --JIN}
%
Here, each predictive model is trained with a qualified maximal subset with its missing rate smaller than $T_{max-missing}$. If the missing rate of the original data is significantly larger than $T_{max-missing}$, the qualified maximal subsets could be much smaller subset of the original data. Therefore, a predictive model can successfully capture the local but not the global properties of the original data. 
Directly using these predictive model individually may not result in optimal results. Meanwhile, for the records in the external validation set, they may differ regarding from which distributions the records are drawn, indicating that we may not obtain the best performance by aggregating all the pre-trained models. Hence, in the ensemble prediction stage, we develop a novel strategy to select pre-trained predictive models according to data representation and ensemble them for external validation.
% subsets are nearly complete -- meaning the missing rates of the qualified maximal subsets are smaller than $T_{max-missing}$

\subsubsection{Constructing Support Set}
To estimate the distribution of the external validation records, a support set is generated. Mathematically, the support set $SS^{k, N_f}$ is generated by randomly select $k$ rows from the original dataset $I$. Similar to $I$, $SS$ may have a significant missingness.  

For $SS$, a binary missingness matrix $MissingSS$ is obtained using the same method described in Section~\ref{sec:Missingness}.

\subsubsection{Measuring Patient Similarity}
For each external validation records, we measure the similarities between it and all the records in the support set $SS$ pair-wisely. Top $k_1$ similar records in $SS$ are selected. ELMV assigns a set of dedicated pre-trained models to each external validation record by selecting all the pre-trained models that can successfully predict at least $k_2$ top records ($k_2 <= k_1$). Both $k_1$ and $k_2$ is a user defined parameter. 

 %\textcolor{red}{Updates Above---------------------------Lucas}

%is associated with $k_1$ predictive models, we can select at most $k_1\times k_2$ pre-trained predictive models for the given test record.
%Meanwhile, for each record in $SS$, we select top-$k_1$ pre-trained models, 

Formally, the similarity between a external validation record and all records in the support set $SS$ is defined in Equation~\ref{sim_equ}:
\begin{equation}
\label{sim_equ}
  Sim = W_F*Softmax(-Dist\_F) + W_M*Softmax(-Dist\_M)
\end{equation}

\noindent where $Sim \in \mathbb{R}^{(1 \times N_f)}$ represents similarity between each individual validation record and all the records in the support set, $Dist\_F \in \mathbb{R}^{(1 \times N_f)}$ represents the Euclidean distance between the corresponding feature vectors. $Dist\_M \in \mathbb{R}^{(1 \times N_f)}$ represents the Hamming distance between the missingness indicator vectors $MissingI^{p}$, and the overall similarity score is a weighted sum of the two distances normalized using softmax. Here, weights $W_F$ and $W_M$ are user adjustable parameters. Larger $W_F$ indicates ELMV pays more attention on feature vector, and similarly, larger $W_M$ indicates the missingness vector is more important than the others. %The larger the similarity score, there is a higher probability that the two compared records are similar. Then the top $K$ similar support records for each external validation set is determined.

\subsubsection{Ensemble Prediction}
Finally, we select multiple pre-trained predictive models and aggregate them by adopting the ensemble prediction approach. 
The model selection procedure can be described as a multi-objective optimization problem that considers the following objectives: the model prediction performance on support records similar to the target external validation records, the model performance on all records in the support set, the model cross-validation performance such as accuracy, precision, recall, and F1, as well as the characteristics of the subset that is used to train the model including the number of features, the number of patients, and the missing rate. 
%\textcolor{red}{Updates Above---------------------------Lucas}

Given a list of model selection criterion $\{C^1,C^2, ... C^n\}$ and a list of candidate models $\{M_1,M_2, ... M_m\} \in M$, let $TBest^{C}_{M}$ be a binary vector indicating whether model $M$ performs the best under criteria $C$. Mathematically,
\begin{equation}
\label{select_func}
   TBest^{C}_{M}=
    \begin{cases}
      1, & \text{if}\ C^{i}_{M_j} = MAX(C^{i})\\
      0, & \text{otherwise}
    \end{cases}
\end{equation}

A pre-trained model is selected if and only if it performs the best on at least one criterion formulated in Equation~\ref{select_func2} or the overall performance in all criterion is the highest (see  Equation~\ref{select_func3}). The number/type of the objectives are user adjustable. %\textcolor{red}{Updates Above---------------------------Lucas}

\begin{equation}
\exists{C}: \in  TBest^{C}_{M} =1
\label{select_func2}
\end{equation}

\begin{equation}
\label{select_func3}
\argmax_{M} \sum_{i=1}^{n} TBest^{C_{i}}_{M}
\end{equation}

In the last step, the final prediction for each record in the external validation set can be obtain by integrating all the selected models. For the demonstration purpose, a majority voting of all the selected models is used here, which can be replaced with other ensemble learning approaches with a simple modification.   

\subsection{ELMV Stage 3: Critical Feature Identification}
Each predictive model trained with a qualified maximal subset produces its own critical features in its local context. In order to identify the critical features of the entire data, we repeatedly apply the leave-one-out cross validation (LOOCV)~\cite{LOOCV_Sammut} on each qualified maximal subset. Finally, we aggregate the most critical features of each predictive model using a weighted voting mechanism. 
The critical feature identification process is shown in Figure~\ref{feature_p}. Through this process, domain experts can examine the validity and reliability of ELMV by checking whether the critical features found is reasonable under both the local and global context. 

In the weighted voting process, the weight of a critical feature is determined by three factors, i.e. the local LOOCV performance of the pre-trained predictive model, missing rate of the qualified maximal subset used to train the predictive model, and local feature importance.

Generally speaking, the higher the local LOOCV performance, the more weight is put on the features found by that predictive model. 
Specifically, for each predictive model, the top-$k_3$ critical features are determined using the feature importance. And then, all the top-$k_3$ local critical features of every predictive model with a similar missing rate are sorted and ranked. %(e.g. Rank all local critical features in $5\%$ category.) 
The feature ranking is based on the ratio between the number of times a given feature being selected and the number of times it is available. Given the ranked feature list, we select top-$k_4$ critical features using weighted voting where weights are determined by the averaged local LOOCV model performance.

\section{Experimental Results}
We tested the performance of ELMV on multiple simulation datasets as well as a real-world EHR dataset. XGBoost was used as the base predictive model in both tests. For performance comparison, we obtained the performance of three models: 1) to impute data with the mean imputation and to train the imputed data with XGBoost, 2) to impute data with MICE~\cite{JSSv045i03} and to train the imputed data with XGBoost, and 3) to train XGBoost without using data imputation. 

\subsection{Data Preprocessing}
\subsubsection{Simulation Data}
To simulate EHR data with a significant missing rate, we selected a complete dataset and based on which, constructed multiple simulation datasets with a wide range of missing rates. On the simulation data, we test whether adopting ELMV is able to achieve performance comparable to that of a predictive model trained on the complete dataset. %\textcolor{red}{Updates Above "adopting ELMV on simulation data"-------Lucas}
%
%on simulation data is to evaluate ELMV frame work on prediction for dataset that has wider range of missing proportion regardless of the domain of the dataset. 
%
The complete dataset obtained was the widely used IRIS dataset for machine learning educational uses from the UCI repository~\cite{Dua:2019}. The IRIS data consists of four features, 150 records, and three outcome labels. The LOOCV accuracy of XGBoost on the IRIS data is 0.97. %, which indicates that there is at least one critical feature in the IRIS data. 
In total, 22 simulation data were constructed using the IRIS data, each having 40 features and 150 records, while the missing rate varying from $5\%$ to $70\%$. 

%\subsubsection{Simulation Data Preprocessing}
%We first confirmed that the original dataset can achieve a LOOCV classification performance of 0.97 which indicates the original dataset is valid for simulation purpose. 

All the simulation datasets were constructed similarly, except that different missing rates were used. 
First, using each of the original features in the IRIS data, we generated nine new features with different noise rates. Here, different noise rates were used to test whether the model can identify and retain high quality features while discarding low quality features. Finally, we randomly removed $5\%$ to $70\%$ entries from every simulation dataset. With this, we simulate the situation that important features in the EHR data, if missing, are likely to be retained by auxiliary features. %The simulation process is done by randomly sample missing values for each feature at different missing percentage.

\subsubsection{Real World Healthcare Data}
The real-world EHR data we used was adopted from a follow-up study of 240 type 2 diabetes (T2DM) patients who went through the Laparoscopic Roux-en-Y Gastric Bypass (LRYGB) surgery~\cite{alexandrides2007resolution}. The LRYGB dataset was collected in the Shanghai Jiaotong University Affiliated 6th People’s Hospital. The data were de-identified before use.  

The LRYGB dataset consists of 79 variables including HbA1c and the other 78 biomedical variables collected at six different time points, i.e. before the LRYGB surgery, 3-month, 6-month, 12-month, 24-month, and 36-month after the surgery. In total, 240 T2DM patients participated the study. %Due to patient drop-out and occasionally mis-recorded measurements, the average missing rates of the LRYGB data at each time point range from $3\%$ to $56\%$. 
%the number 3to56 is the average missing rate for each feature at each time point and it is for 214 pts after labelling, I  need to get the number for 240pts. %
%
Among all the 78 biomedical variables, 24 of them, such as CysC, weight index, and direct bilirubin, were pre-selected based on domain knowledge for further studies. 

\begin{figure}[bt!]
\centering
\includegraphics[width=0.8\linewidth]{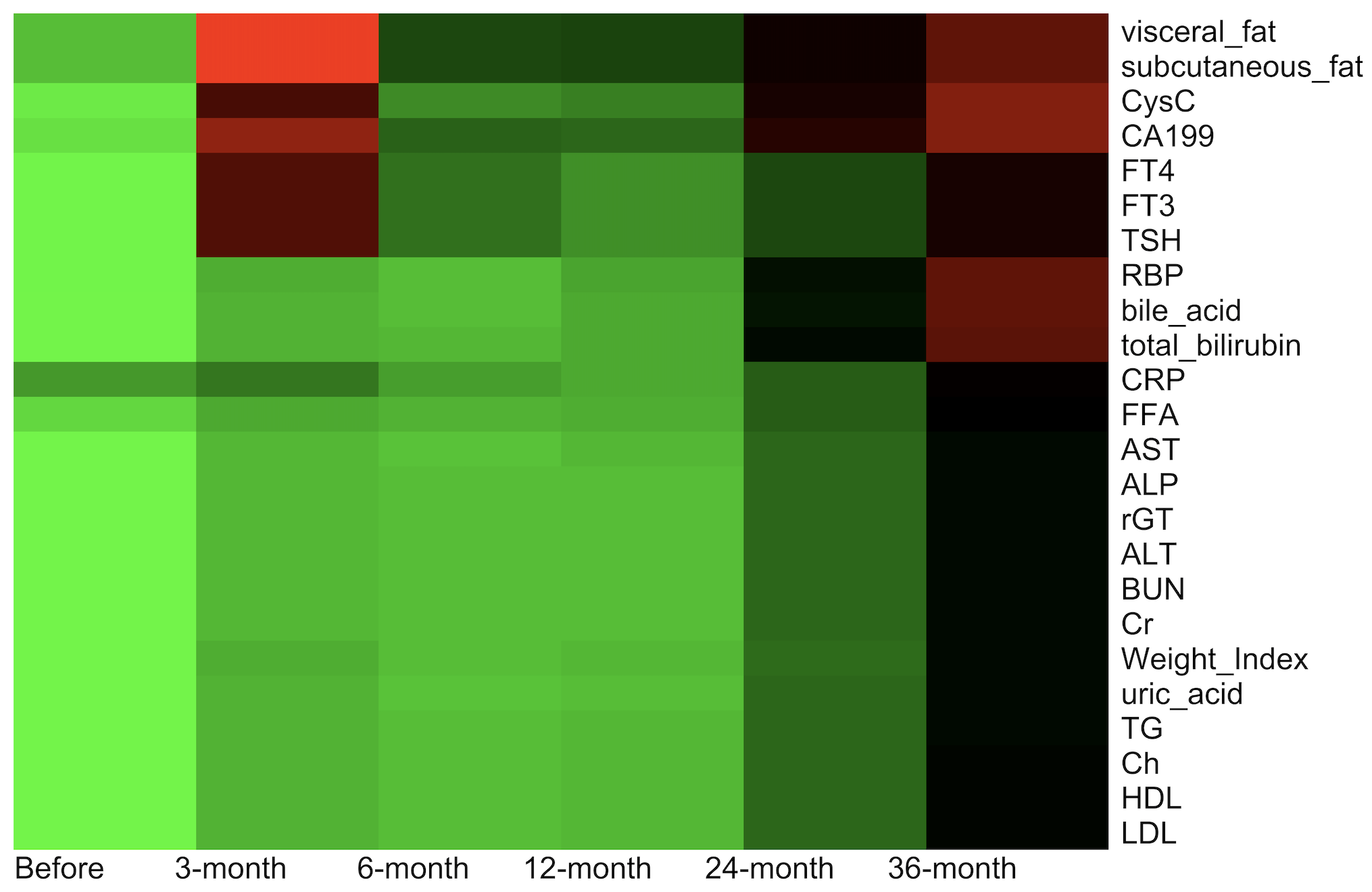}
\caption{In the LRYGB follow-up study, the distribution of the missing values of all the 24 variables at six time points. In general, more values are missing towards the end of the follow-up study. Red indicates higher missing ratio towards $100\%$, green is for lower missing ratio towards $0\%$, and black indicates $50\%$ missing ratio.}
\label{missingR_f1}
\end{figure}

%\textbf{Labelling.} 
The types of HbA1c trajectories were determined using clustering followed by manual curation. Specifically, we adopted the reversed K-nearest neighbor (rKNN)~\cite{Papadias2009} to remove outliers and the agglomerative hierarchical clustering with ward~\cite{WardMethod} to separate all the patients into nine clusters. Then Elbow method was used to determine the optimal number of clusters, on which the decreasing rate of With-in-Sum-of-Squares (WSS) was the slowest. %Nine clusters are obtained.
Furthermore, two clinicians checked the obtained clusters independently and defined five HbA1c trajectory labels. %In addition, clinicians manually examined the incomplete data and identified an extra HbA1c trajectory cluster resulting in six types of HbA1c trajectories labels.
In summary, after semi-automatic labelling, the LRYGB follow-up data consists of 214 patients, 24 features, and six labels. The missingness of all the features of the LRYGB data is shown in Figure~\ref{missingR_f1}. The missing ratio at every time point is 
$3\%$, $33\%$, $18\%$, $18\%$, $37\%$, and $56\%$ respectively. Clearly, patient drop out is a main issue that resulted in large missing rates at later time points. Use this real-world data, we can test ELMV at the non-random missing data situation. Specifically, we evaluated ELMV by testing whether it can identify critical features that can be used to predict the temporal trajectory of HbA1c.

%We done (1/fxp) , WE think : (1/f*p*6)
%\subsubsection{Real World Healthcare Data Preprocessing}
As part of the data preprocessing, we imputed a small portion of the missing values using domain knowledge and simple statistics such as linear interpolation. Also, we copied the $6^{th}$ month values to the $3^{rd}$ month, if the $3^{rd}$ month values were missing. We removed patients whose HbA1c values at both $3^{rd}$ month and $6^{th}$ month are missing. %\textcolor{blue}{We only remove patients who has both missing for HbA1c, not for all features}% Then we did linear interpolation for all other missing entries if there exist non-missing values both before and after the missing entry. 
After this step, the LRYGB follow-up data consists of 202 patients, 24 features, and the overall missing rate of the LRYGB data was reduced. For example, the missing rates at 24-month and 36-month have been effectively reduced from 37\% to 25\% and from 56\% to 48\% respectively. But still, the high missing rate towards the end of the T2DM follow-up  study prevent us from using any predictive models directly.  

%3\%, 7\%, 8\%, 8\%, 25\%, 48\%. 
%
%With some of the features have high missing ratio at 3 year after operation as an exception. %\textcolor{blue}{Not sure which figure to keep}

%For this study, we only focuses on identifying the 24 features were thought to be HbA1c-unrelated but are actually critical for predicting HbA1c trajectory in the first 3 years of the Roux-en-Y gastric bypass operation.  

 %In addition, the missing trend for all features are shown in Figure ~\ref{missingR_f2}. 

\subsection{Experimental results}
We applied ELMV, as well as three baseline algorithms, i.e. mean imputation, MICE, and XGBoost without data imputation, on both the simulation data and the real-world EHR data. For model performance comparison, conventional classification metric were used, including accuracy, precision, recall, and F-1. Specifically, domain experts manually reviewed the critical features identified on the EHR data assessing whether they can be used to predict the temporal trajectory of HbA1c.

\subsubsection{Prediction Performance on Simulation Data}
\begin{figure}
\centering
\includegraphics[width=0.5\textwidth]{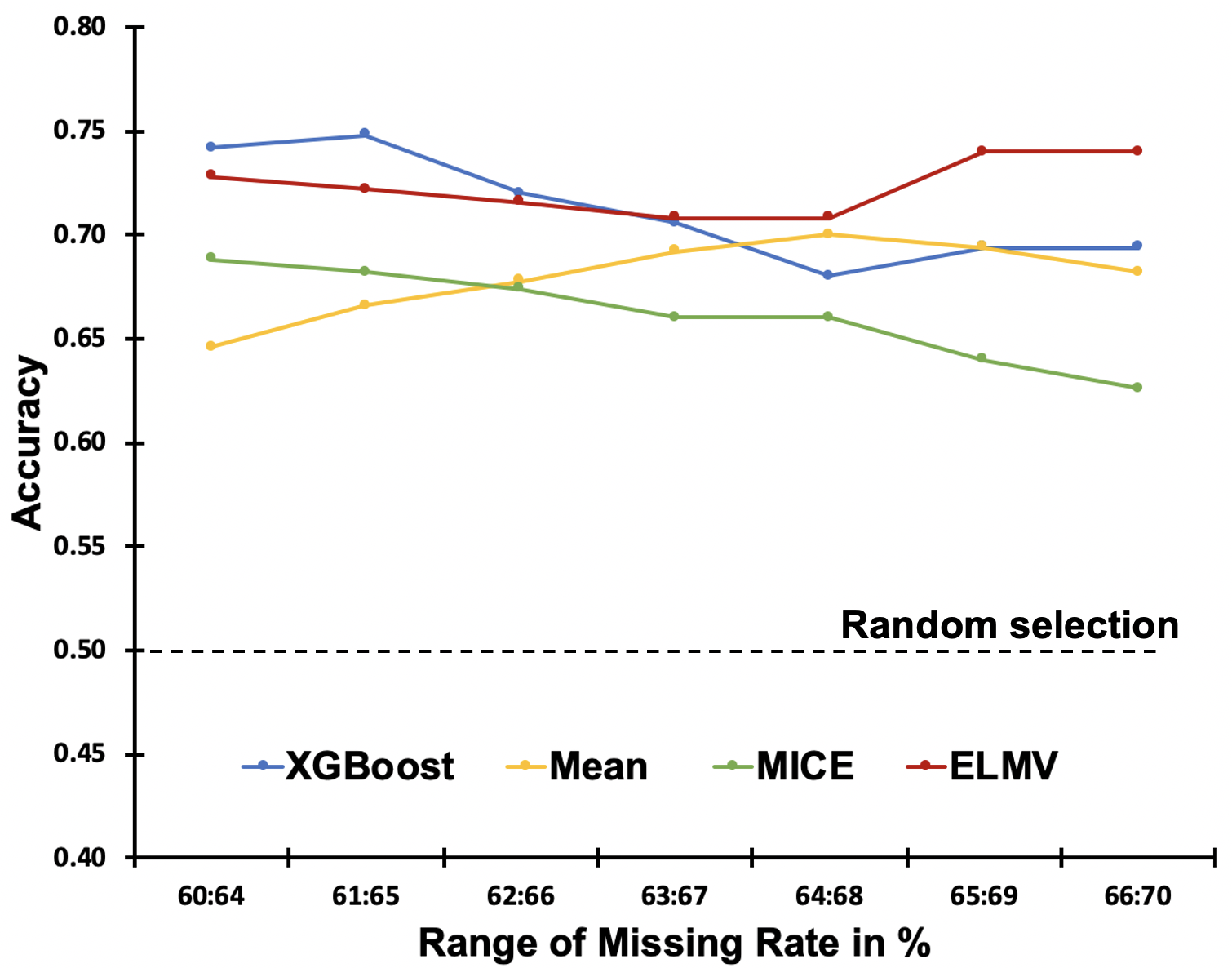}
\caption{The moving average of accuracy of ELMV, XGBoost, and two data imputation methods on the simulation data with missing rate increasing from 60\% to 70\%.}
\label{sim_performance_trend}
\end{figure}

On all the simulation IRIS datasets with their missing rates ranging from $5\%$ to $70\%$, the performance of ELMV, mean imputation, MICE, and XGBoost without data imputation was systematically compared. Table~\ref{sim_perf_table} compares model prediction accuracy of the four models on the simulation datasets. When the missing rate was low (5\% to 20\%), all the models can achieve nearly perfect performance (accuracy $\geq 0.93$). However, if the missing rate was in the range of 60\% and 70\%, the accuracy of all the models to compare was reduced significantly below 75\% no matter how the missing values were handled. ELMV still can maintain its accuracy above 75\%. A moving average of accuracy on finer granularity of missing rates shown in Figure~\ref{sim_performance_trend} reveals that ELMV is not affected by the high missing rates as bad as the other three models. The performance trends in Figure~\ref{sim_performance_trend} suggest that, ELMV can achieve the best accuracy towards larger missing rates steadily when the missing rate was increased, which XGBoost achieved the best performance if the missing rate was relatively low. MICE had the overall lowest accuracy and its accuracy trend dropped steadily when the missing rate was increased. Surprisingly, the mean imputation had a relative stable performance, probably because the missing values were removed completely randomly. However, the mean imputation never performed the best on any dataset. Both mean imputation and MICE have lower accuracy than XGBoost, indicating that the two data imputation methods tested failed to reinforce XGBoost's ability to handle missing values. %whereas ELMV increase the ability of handling missing values of XGBoost.

\begin{figure}
\centering
\includegraphics[width=0.5\textwidth]{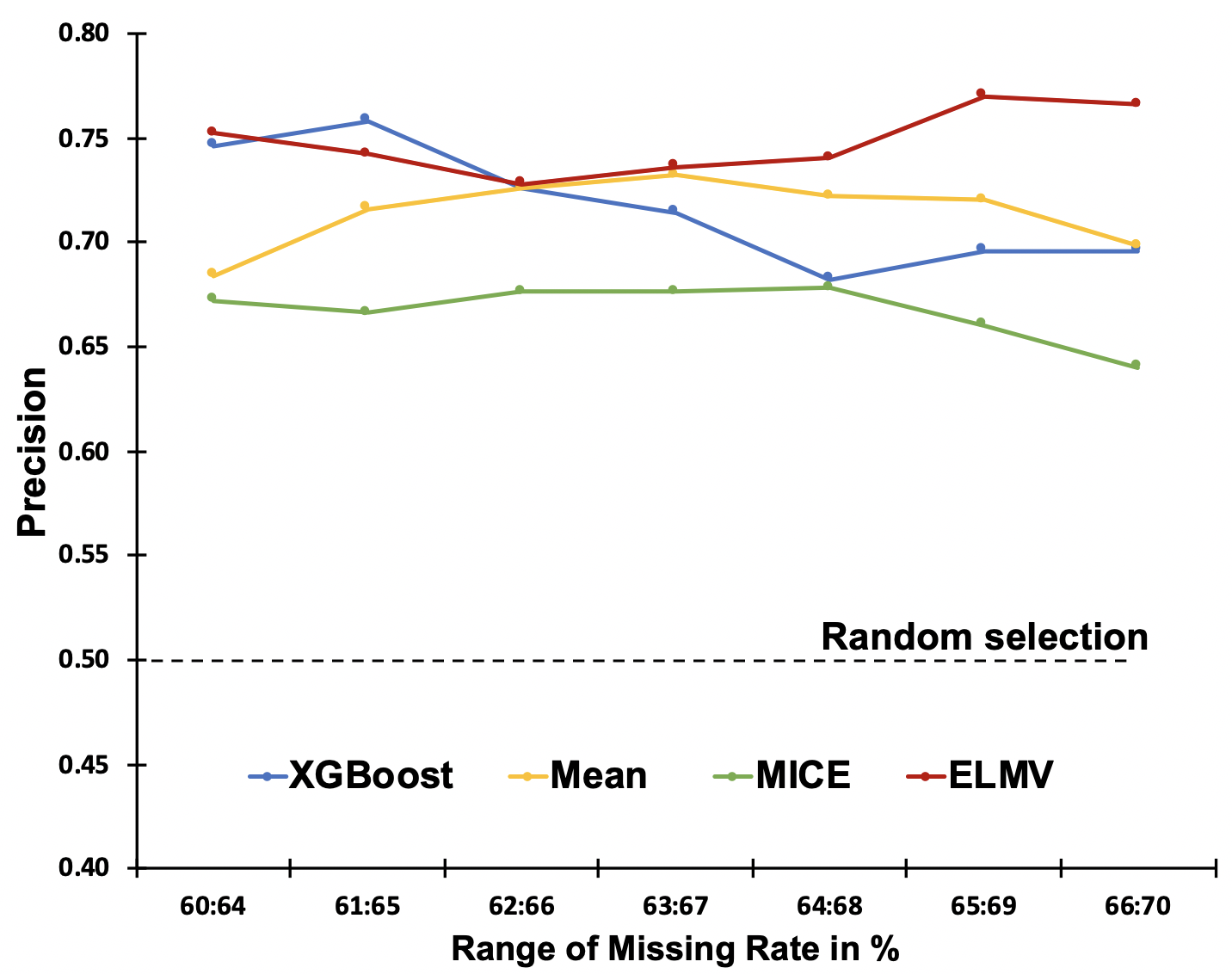}
\caption{The moving average of precision of ELMV, XGBoost, and two data imputation methods on the simulation data with missing rate increasing from 60\% to 70\%.}
\label{sim_precision_trend}
\end{figure}

\begin{figure}
\centering
\includegraphics[width=0.5\textwidth]{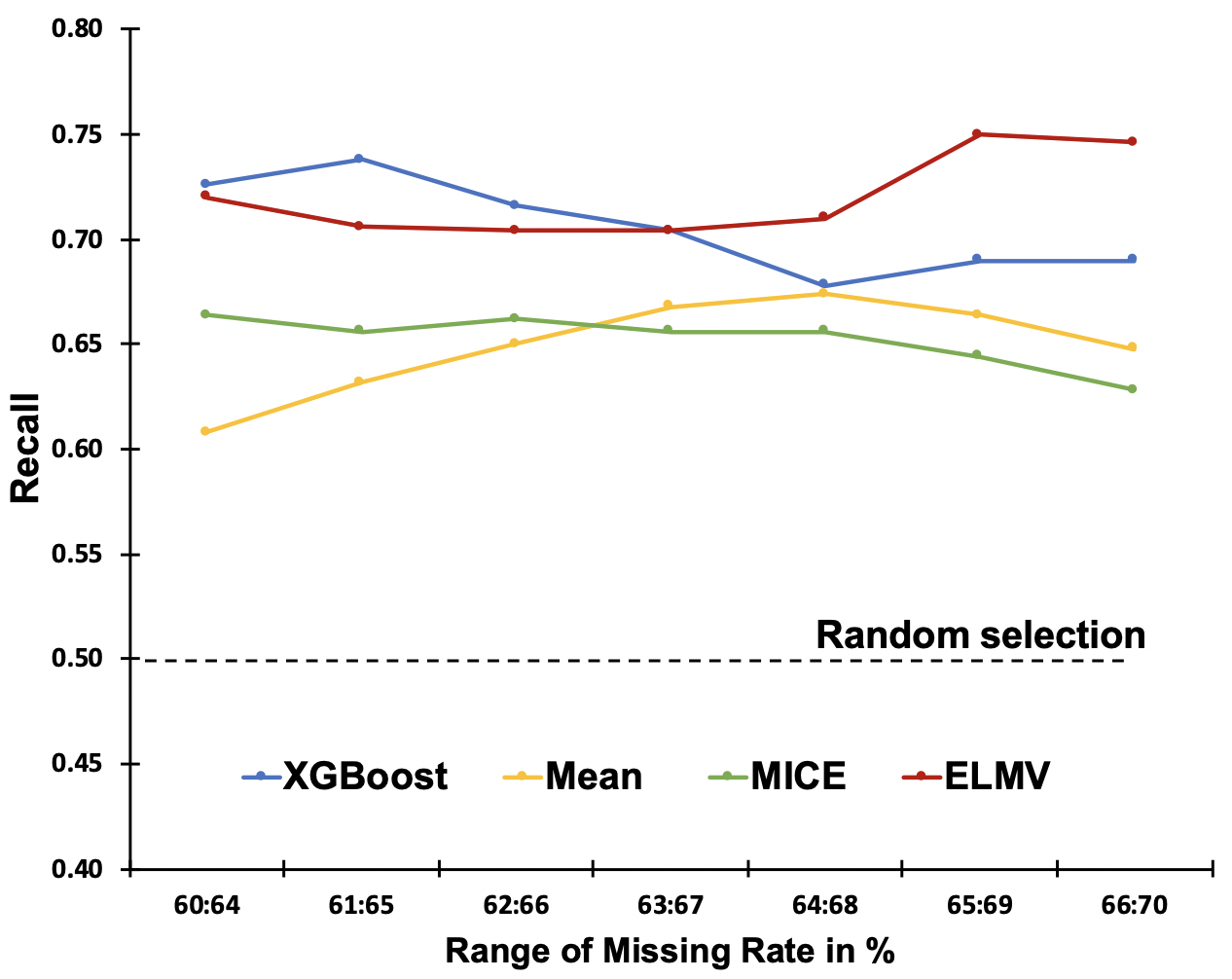}
\caption{The moving average of recall of ELMV, XGBoost, and two data imputation methods on the simulation data with missing rate increasing from 60\% to 70\%.}
\label{sim_recall_trend}
\end{figure}

\begin{figure}
\centering
\includegraphics[width=0.5\textwidth]{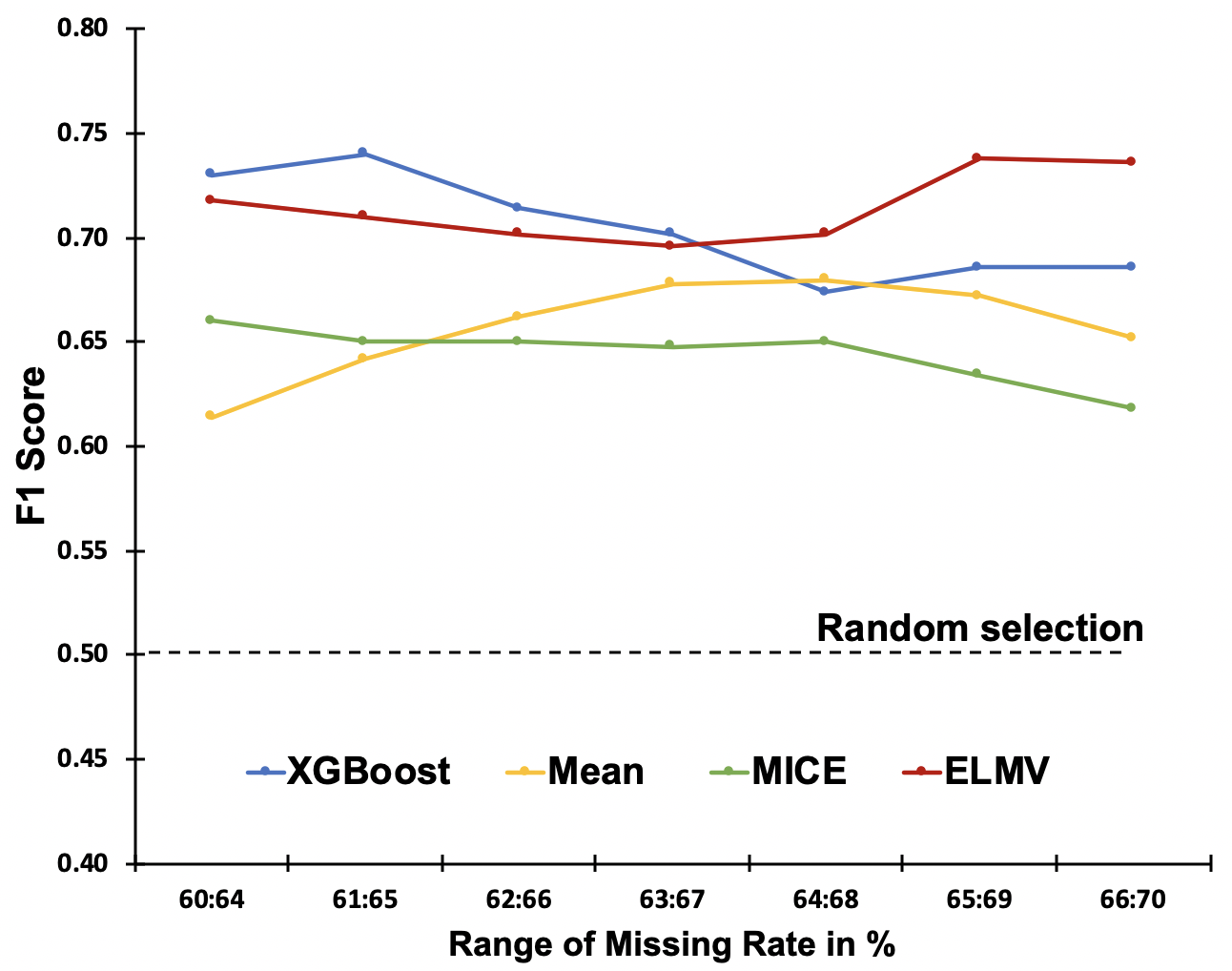}
\caption{The moving average of F-1 Scores of ELMV, XGBoost, and two data imputation methods on the simulation data with missing rate increasing from 60\% to 70\%.}
\label{sim_f1_trend}
\end{figure}

The averaged precision, recall, and F-1 are reported in Table~\ref{sim_precision_table}, Table ~\ref{sim_Recall_table}, and Table~\ref{sim_F1_table}, respectively. Similarly,
ELMV achieved the best performance in all but one case when the missing rate was high. A moving average of precision, recall, and F-1 on finer granularity of missing rates shown in Figure~\ref{sim_precision_trend}, Figure~\ref{sim_recall_trend}, and Figure~\ref{sim_f1_trend} indicate that ELMV achieved overall the best performance.   

%computed over all class labels. Performance of average precision,recall and F1 compared to baseline methods are shown in Table  respectively. Plus, the performance in accuracy trend is shown in Figure ~\ref{sim_performance_trend}.

%The results shows that our ELMV framework has a better performance in terms of all metrics: average accuracy, precision, recall and F1 over all simulation data in different missingness proportion. Especially, in the range of $67\%$ to $73\%$, ELMV has the best performance in terms of Accuracy. 
%All methods performed similarly well when the missing rate was small. However, when the missing rate was in the range of $60\%$ to $70\%$, the performance varied. 

\begin{table}[ht]
\centering
\caption{Averaged accuracy of ELMV, XGBoost, and two data imputation methods on the simulation data with low (above the horizontal line) or high missing rates (below the horizontal line).}
\label{sim_perf_table}
\begin{tabular}{ccccc}
\hline
\textbf{Missing} & \textbf{XGBoost} & \textbf{Mean} & \textbf{MICE} & \textbf{ELMV} \\ 
\textbf{Rate} & \textbf{} & \textbf{Imputation} & \textbf{Imputation} & \textbf{} \\ \hline
5\%   & 0.97  & 0.97 & 0.97          & 0.97          \\
10\%  & 1.00  & 0.97 & 0.93          & 1.00          \\
20\%  & 0.97  & 0.97 & 0.97          & 0.97          \\
%30\%  & 0.87  & 0.90 & 0.93          & 0.83          \\
\hline
%40\%  & 0.93  & 0.83 & 0.87          & 0.90          \\
%50\%  & 0.83  & 0.83 & 0.87          & 0.90          \\\hline
60\%  & 0.67  & 0.63 & 0.73          & \textbf{0.80}          \\
%61\%  & 0.77  & 0.67 & 0.77          & \textbf{0.80}          \\
%62\%  & \textbf{0.77}  & 0.60 & 0.60          & 0.67          \\
%63\%  & \textbf{0.80}  & 0.63 & 0.77          & 0.70          \\
%64\%  & \textbf{0.70}  & 0.70 & 0.57          & 0.67          \\
65\%  & 0.70  & 0.73 & 0.70          & \textbf{0.77}          \\
%66\%  & 0.63  & 0.73 & 0.73          & \textbf{0.77}          \\
%67\%  & \textbf{0.70}  & 0.67 & 0.53          & 0.63          \\
%68\%  & 0.67  & 0.67 & \textbf{0.77}          & 0.70          \\
%69\%  & 0.77  & 0.67 & 0.47          & \textbf{0.83}          \\
70\%  & 0.70  & 0.67 & 0.63          & \textbf{0.77}          \\
%71  & 0.80  & 0.70 & 0.53          & 0.77          \\
%72  & 0.63  & 0.70 & 0.47          & 0.77          \\
%73  & 0.70  & 0.67 & 0.50          & 0.73          \\
%74  & 0.60  & 0.63 & 0.33          & 0.50          \\
%75  & 0.63  & 0.67 & 0.60          & 0.63          \\
%Average      & 0.76  & 0.74        & 0.69          & 0.78          \\ 
\hline
\end{tabular}
\end{table}

\begin{table}[ht]
\centering
\caption{Averaged precision of ELMV, XGBoost, and two data imputation methods on the simulation data with low (above the horizontal line) or high missing rates (below the horizontal line).}
\label{sim_precision_table}
\begin{tabular}{ccccc}
\hline
\textbf{Missing} & \textbf{XGBoost} & \textbf{Mean} & \textbf{MICE} & \textbf{ELMV} \\ 
\textbf{Rate ($\%$)} & \textbf{} & \textbf{Imputation} & \textbf{Imputation} & \textbf{} \\ \hline
5  & 0.97 & 0.97 & 0.97 & 0.97 \\
10 & 1.00 & 0.97 & 0.95 & 1.00 \\
20 & 0.97 & 0.97 & 0.97 & 0.97 \\
\hline
%30 & 0.86 & 0.93 & 0.95 & 0.83 \\
%40 & 0.93 & 0.83 & 0.90 & 0.92 \\
%50 & 0.82 & 0.88 & 0.91 & 0.90 \\
60 & 0.65 & 0.64 & 0.75 & \textbf{0.83} \\
%61 & 0.79 & 0.71 & 0.74 & \textbf{0.86} \\
%62 & \textbf{0.77} & 0.65 & 0.56 & 0.71 \\
%63 & \textbf{0.81} & 0.69 & 0.75 & 0.68 \\
%64 & 0.71 & 0.73 & 0.56 & 0.68 \\
65 & 0.71 & \textbf{0.80} & 0.72 & 0.78 \\
%66 & 0.63 & 0.76 & \textbf{0.79} & \textbf{0.79} \\
%67 & \textbf{0.71} & 0.68 & 0.56 & 0.75 \\
%68 & 0.65 & 0.64 & 0.76 & 0.70 \\
%69 & 0.78 & 0.72 & 0.47 & \textbf{0.83} \\
70 & 0.71 & 0.69 & 0.62 & \textbf{0.76} \\
%71 & 0.82 & 0.71 & 0.52 & 0.79 \\
%72 & 0.64 & 0.78 & 0.47 & 0.76 \\
%73 & 0.68 & 0.68 & 0.47 & 0.72 \\
%74 & 0.63 & 0.63 & 0.32 & 0.55 \\
%75 & 0.68 & 0.66 & 0.60 & 0.69 \\
\hline
\end{tabular}
\end{table}

\begin{table}[ht]
\centering
\caption{Averaged recall of ELMV, XGBoost, and two data imputation methods on the simulation data with low (above the horizontal line) or high missing rates (below the horizontal line).}
\label{sim_Recall_table}
\begin{tabular}{ccccc}
\hline
\textbf{Missing} & \textbf{XGBoost} & \textbf{Mean} & \textbf{MICE} & \textbf{ELMV} \\ 
\textbf{Rate ($\%$)} & \textbf{} & \textbf{Imputation} & \textbf{Imputation} & \textbf{} \\ \hline
5    & 0.95 & 0.95 & 0.95 & 0.95 \\
10   & 1.00 & 0.95 & 0.90 & 1.00 \\
20   & 0.95 & 0.95 & 0.95 & 0.95 \\
\hline
%30   & 0.85 & 0.87 & 0.90 & 0.80 \\
%40   & 0.93 & 0.82 & 0.83 & 0.88 \\
%50   & 0.82 & 0.78 & 0.81 & 0.92 \\
60   & 0.64 & 0.59 & 0.73 & \textbf{0.83} \\
%61   & 0.74 & 0.62 & 0.73 & \textbf{0.79} \\
%62   & \textbf{0.76} & 0.56 & 0.57 & 0.67 \\
%63   & \textbf{0.79} & 0.59 & 0.76 & 0.67 \\
%64   & \textbf{0.70} & 0.68 & 0.53 & 0.64 \\
65   & 0.70 & 0.71 & 0.69 & \textbf{0.76} \\
%66   & 0.63 & 0.71 & 0.76 & \textbf{0.78} \\
%67   & \textbf{0.70} & 0.65 & 0.54 & 0.67 \\
%68   & 0.66 & 0.62 & \textbf{0.76} & 0.70 \\
%69   & 0.76 & 0.63 & 0.47 & \textbf{0.84} \\
70   & 0.70 & 0.63 & 0.61 & \textbf{0.74} \\
%71   & 0.81 & 0.66 & 0.53 & 0.78 \\
%72   & 0.64 & 0.68 & 0.50 & 0.74 \\
%73   & 0.68 & 0.65 & 0.48 & 0.72 \\
%74   & 0.61 & 0.63 & 0.30 & 0.51 \\
%75   & 0.64 & 0.65 & 0.58 & 0.67 \\
\hline
\end{tabular}
\end{table}

\begin{table}[ht]
\centering
\caption{Averaged F-1 of ELMV, XGBoost, and two data imputation methods on the simulation data with low (above the horizontal line) or high missing rates (below the horizontal line).}
\label{sim_F1_table}
\begin{tabular}{ccccc}
\hline
\textbf{Missing} & \textbf{XGBoost} & \textbf{Mean} & \textbf{MICE} & \textbf{ELMV} \\ 
\textbf{Rate ($\%$)} & \textbf{} & \textbf{Imputation} & \textbf{Imputation} & \textbf{} \\ \hline
5  & 0.96 & 0.96 & 0.96 & 0.96 \\
10 & 1.00 & 0.96 & 0.92 & 1.00 \\
20 & 0.96 & 0.96 & 0.96 & 0.96 \\
\hline
%30 & 0.85 & 0.89 & 0.92 & 0.81 \\
%40 & 0.93 & 0.82 & 0.85 & 0.89 \\
%50 & 0.82 & 0.79 & 0.82 & 0.90 \\
60 & 0.64 & 0.59 & 0.74 & \textbf{0.80} \\
%61 & 0.76 & 0.62 & 0.72 & \textbf{0.80} \\
%62 & \textbf{0.76} & 0.57 & 0.55 & 0.67 \\
%63 & \textbf{0.79} & 0.60 & 0.75 & 0.67 \\
%64 & \textbf{0.70} & 0.69 & 0.54 & 0.65 \\
65 & 0.69 & 0.73 & 0.69 & \textbf{0.76} \\
%66 & 0.63 & 0.72 & 0.72 & \textbf{0.76} \\
%67 & \textbf{0.70} & 0.65 & 0.54 & 0.64 \\
%68 & 0.65 & 0.61 & \textbf{0.76} & 0.70 \\
%69 & \textbf{0.76} & 0.65 & 0.46 & 0.83 \\
70 & 0.69 & 0.63 & 0.61 & \textbf{0.75} \\
%71 & 0.79 & 0.67 & 0.52 & 0.76 \\
%72 & 0.63 & 0.71 & 0.48 & 0.75 \\
%73 & 0.68 & 0.65 & 0.47 & 0.72 \\
%74 & 0.60 & 0.62 & 0.46 & 0.50 \\
%75 & 0.62 & 0.65 & 0.59 & 0.63

\hline
\end{tabular}
\end{table}

\subsubsection{Feature Selection on Real World EHR Data}
\begin{figure}[h!]
\centering
\includegraphics[width=0.5\textwidth]{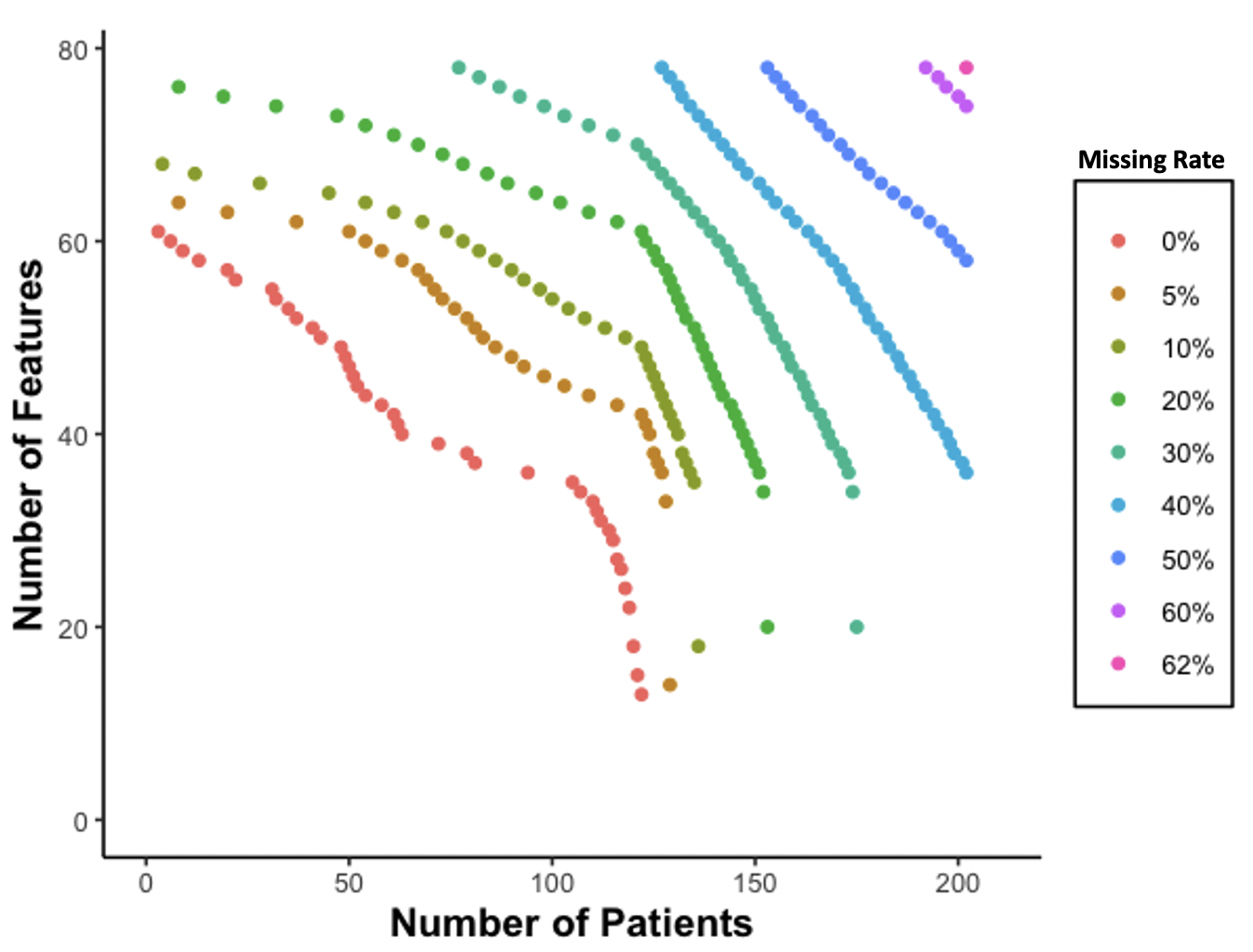}
\caption{The distribution of all the maximal subset of the original LRYGB data with 78 features and 202 T2DM patients. Every point represents a maximal subset with $x$ number of patients and $y$ number of features. Color indicates different missing rates.}
\label{missingratio_p}
\end{figure}

We applied ELMV on the LRYGB dataset aiming at identifying critical features for HbA1c trajectory prediction. 
All the qualified maximal subsets of the LRYGB data, which has 78 features and 202 T2DM patients are generated are shown in Figure~\ref{missingratio_p}. Every point in the figure represents a qualified maximal subset of the the LRYGB dataset. X-axis indicates the number of patients and y-axis indicates the number of features of the qualified maximal subset. 

All the points with the same color have a similar missing rate. We generated all qualified maximal subsets for the LRYGB data including all 78 features so that any combinations of features of interest can be evaluated in critical feature identification stage of ELMV. 
Since the goal of this experiment is to identify the critical features among 24 pre-selected features, we only used the qualified maximal subsets that contains these 24 pre-selected features for validation purposes.

Early stage biomarkers, such as serum Ca2+ and cholesterol level measured at 3-month ~\cite{berridge2000versatility,ammala1993calcium}, were found by ELMV to be critical for predicting HbA1c trajectory in the first 3 years after the LRYGB surgery. 

In addition, the overall accuracy of ELMV is 0.93, significantly higher than that of XGBoost (0.63), Mean imputation (0.30), and MICE (0.28). 
The performance of ELMV on all the qualified maximal subsets with missing rate ranging from 0\% to 30\% is shown in Table~\ref{tab1}. The table indicates that ELMV can maintain its accuracy above $90\%$ and is not significantly affected by the missing rates of the qualified maximal subsets.

%We further tested ELMV using records from 33 new patients. The external text shows that among the 33 new patients, the HbA1c trajectories of 26 patients were correctly classified. 
%\textcolor{blue}{I am not sure if we should add the sentences above. 1st, we do not test the ELMV for prediction on these 33 new patietns, we only check if the critical features found by ELMV can be used to correctly classify them. 2nd, we did not use the 24 features mentioned in this MS }

%Considering that this is an external validation test, it is nor surprise that ELMV  %In addition, the LOOCV performance from ELMV archives an accuracy of 0.95 which is substantially higher than other baseline models. 
%The performance comparison between ELMV framework with baseline algorithms is shown in Table~\ref{tab2}

\begin{table}[h!]
\centering
\caption{Performance of qualified maximal subsets  of the LRYGB data with different missing rates.}
\label{tab1}
  \begin{tabular}{cc}
  \toprule
   \textbf{Missing Rate ($\%$)} & \textbf{Accuracy} \\ 
   \midrule
    $0$  & 0.90   \\ 
    $5$   & 0.95   \\ 
    $10$   & 0.94   \\
    $20$   & 0.94   \\
    $30$   & 0.93   \\
%    $40\%$   & 0.94   \\
%    $50\%$   & 0.92   \\
%    $60\%$   &0.82   \\ 
%    Weighted Voting   &0.95   \\ 
\hline
    Average & 0.93 \\
   \bottomrule 
\end{tabular}
\end{table}

%\begin{table}
%\caption{Performance comparison between ELMV Framework and Baseline Algorithms}
%\label{tab2}
%  \begin{tabular}{cc}
%  \toprule
%   \textbf{Method} & \textbf{Accuracy} \\ 
%   \midrule
%     ELMV    &0.95   \\ 
%     XGBoost   &0.63  \\ 
%     Mean imputation   &0.30   \\
%     MICE    &0.28    \\
%\bottomrule 
%\end{tabular}
%\end{table}

\section{Discussion and Conclusion}

\begin{table}[h!]
\centering
\caption{Performance of ELMV and kNN on the LRYGB data.}
\label{sim_copy_comp}
\begin{tabular}{ccc}
    \toprule
   \textbf{Missing Rate (\%)} & \textbf{ELMV} & \textbf{kNN} \\
   \midrule
%    5   & 0.97          & 0.90  \\
%    10  & 1.00          & 0.53  \\
%    20  & 0.97          & 0.33  \\
%    30  & 0.83          & 0.30  \\
%    40  & 0.90          & 0.33  \\
%    50  & 0.90          & 0.43  \\
    60  & 0.80          & 0.53  \\
%    61  & 0.80          & 0.50  \\
%    62  & 0.67          & 0.63  \\
%    63  & 0.70          & 0.50  \\
%    64  & 0.67          & 0.43  \\
    65  & 0.77          & 0.43  \\
%    66  & 0.77          & 0.40  \\
%    67  & 0.63          & 0.40  \\
%    68  & 0.70          & 0.43  \\
%    69  & 0.83          & 0.40  \\
    70  & 0.77          & 0.40  \\
 %   71  & 0.77          & 0.40  \\
 %   72  & 0.77          & 0.30  \\
 %   73  & 0.73          & 0.37  \\
 %   74  & 0.50          & 0.27  \\
 %   75  & 0.63          & 0.40  \\
 %   Average &  0.78     & 0.45      
\bottomrule
\end{tabular}
\end{table}

In this paper, we propose a novel ensemble learning model called ELMV to analyze EHR data with substantial missing values. ELMV outperformed two widely used data imputation methods and an ensemble learning model for patient outcome prediction and critical feature identification in all performance metrics, i.e. accuracy, precision, recall, and F-1. We also demonstrated that ELMV has enough flexibility to take into account data distributions in training and validation in terms of both missingness and actual values. 

In ELMV, a new support set is introduced to estimate the distribution of external validation data and to guide the ensemble learning. %It would be interesting to see how well the extracted data distribution information can be further used in the model training process.  
An interesting question is to what extent the support set can contribute to the ensemble learning, since it is useful only when the external validation data are known. To this end, we compared ELMV with the k-nearest neighbor (kNN) model, which simply assigns each external validation record with the label of most similar records in the support  set. The result shown in table~\ref{sim_copy_comp} indicates that the simple label voting by similar records is unlikely to provide the correct prediction most of time. This experiment further confirms the contribution of the support set in ensemble learning. %to the final prediction is to estimating the distribution of external validation set so as to find the best representative models for final predictions. 

In future work, we plan to optimize the parameter tuning problem of ELMV to further improve its performance. %two concerns by adopting the ideas of Meta-Learning methods where we will try to modify the current ELMV framework for learning the parameters and distribution from the training process, in turn to benefit the prediction processes. 
There are many user-adjustable parameters such as the number of similar records in support set $SS$, the number of top similar records, the number of top pre-trained models, weights for similarity scoring, and the number/type of objectives used in ensemble learning. It would be beneficial if part of these parameters can be learned and auto-tuned during the model training process.

\section{ACKNOWLEDGMENTS}
This project is supported by the Kentucky Lung Cancer Research Program (grant no. KLCR-3048113817) to JL and JC and by the Clinical Retrospective Study of Shanghai Jiaotong University Affiliated 6th People’s Hospital (grant no. YNHG201912) to HZ and JD. Chinese Clinical Trial Registry Number: ChiCTR-ONN-17012895.

\bibliographystyle{unsrt}  
\bibliography{ms.bib}  

\end{document}